\def\year{2019}\relax
\documentclass[letterpaper]{article} 
\usepackage{aaai19}  
\usepackage{times}  
\usepackage{helvet}  
\usepackage{courier}  
\usepackage{url}  
\usepackage{graphicx}  
\usepackage{amsmath}
\usepackage{amssymb}
\usepackage{mathtools}
\usepackage{multirow}
\usepackage{placeins}
\usepackage{booktabs} 
\usepackage{hyperref}
\newcommand{\concat}{%
  \mathbin{{+}\mspace{-8mu}{+}}%
}
\FloatBarrier
\nocopyright
 
\begin{document}
%
\title{MCRM: Mother Compact Recurrent Memory}
\author{Abduallah A. Mohamed \ and Christian Claudel\\
abduallah.mohamed@utexas.edu \\ christian.claudel@utexas.edu}

\maketitle
\begin{abstract}
LSTMs and GRUs are the most common recurrent neural network architectures used to solve temporal sequence problems. The two architectures have differing data flows dealing with a common component called the cell state (also referred to as the memory). We attempt to enhance the memory by presenting a modification that we call the Mother Compact Recurrent Memory (MCRM). MCRMs are a type of a nested LSTM-GRU architecture where the cell state is the GRU hidden state. The concatenation of the forget gate and input gate interactions from the LSTM are considered an input to the GRU cell. Because MCRMs has this type of nesting, MCRMs have a compact memory pattern consisting of neurons that acts explicitly in both long-term and short-term fashions. For some specific tasks, empirical results show that MCRMs outperform previously used architectures.
\end{abstract}

\section{Introduction}
Recurrent neural networks (RNNs) are a class of neural networks that can relate temporal information. They have been widely used for a lot of problems including image caption generation \cite{karpathy2015deep,vinyals2015show,niu2017hierarchical,AAAI1714880}, text to speech generation \cite{fan2014tts,rao2015grapheme,arik2017deep}, object detection and tracking \cite{lu2017online,wolf2017recurrent,kang2017object,yuan2017temporal,tripathi2016context,AAAI1817074}, neural machine translation \cite{bahdanau2014neural,luong2015effective,luong2015stanford,AAAI1714151,AAAI1816788}. RNNs are generally categorized into three well-known architectures, Vanilla RNNs, Long short-term memory (LSTM) proposed by \cite{hochreiter1997long} and Gated recurrent neural networks (GRU) proposed by \cite{GRUSOURCE}. The vanishing and exploding gradients are well-known Vanilla RNNs problems, which GRUs and LSTMs solves. Vanilla RNNs also lack the ability to remember long-term sequences, unlike GRUs and LSTMs. The main difference between LSTMs and GRUs are in the terms of architecture. LSTMs have more control gates than GRUs do. LSTM output is a part of the cell state content unlike GRUs which its cell state is the output. Another difference is that GRUs are clearly computationally inexpensive, unlike LSTMs. Still there is no clear evidence that GRUs is better than LSTMs or not, see for instance the work of \cite{bai2018empirical}.\\

The core idea of LSTMs and GRUs is to control information flow to the cell state, which can be described as the memory, through control gates. Yet the cell state is very simple neural network layer. The core idea of the article is to investigate if the the performance of LSTMs or GRUs can be enhanced by developing a better cell state architecture. We noticed that Nested LSTMs (NLSTM) introduced by \cite{NLSTM} outperforms previous RNN architectures, nonetheless the inner LSTM isn't fully utilized as the cell state is exposed to the outer NLSTM via the output gate. Thus we chose to create a new deep recurrent model that has a GRU unit nested within a LSTM. The GRU is chosen to be inside the LSTM as it fully exposes the hidden state. The GRU represents the cell state of the LSTM. We call this architecture Mother Compact Recurrent Memory (MCRM). The Mother term came from our visualization of the LSTM as a mother that carries the GRU as a fetus. The compact term came from the compact memory pattern that is produced by MCRM which inherits both GRU and LSTM memory behaviors.\\

The MCRMs are positioned as follow:
\begin{enumerate}
\item A novel class of nested RNNs. 
\item A compact memory pattern that support both long and short terms behaviors. 
\item The model is validated using empirical test problems.\\
\end{enumerate}

The rest of this article is organized as follows. Section \ref{secrltwork} reviews the history of RNNs, LSTMs and GRUs and their development highlighting similar approaches. Section \ref{MCRMSECTION} discusses the MCRM model in details, and provides its mathematical model. An experimental validation of MCRM against different recurrent architectures on well-known benchmark recurrent tasks is shown in \ref{experiments-sec}. Section \ref{visualizationsec} shows a visualization of MCRM hidden state and compares these with other RNNs cell states. This demonstrates the compact memory pattern outlined earlier. The MCRM source code is available at: \href{https://github.com/abduallahmohamed/MCRM}{https://github.com/abduallahmohamed/MCRM}.

\section{Related Work} \label{secrltwork}
One of the earliest work in the history of RNNs was by \cite{jordan1997serial}, this work represented an early concept of state in neural networks. It described a recurrent connection with an in-unit loop. It successfully integrated a time series data into a neural network. 
A simpler RNN architecture was proposed by \cite{elman1990finding} which can be called the Vanilla RNN. It simplified the concept of \cite{jordan1997serial} to use a context unit or what can be called a hidden unit removing the in-unit loop that that was previously introduced. The work done by \cite{Lipton15} provides more details about the history of RNNs and its development until it was formalized into the LSTM architecture. Different attempts to improve RNNs itself have been made, including the introduction of an auxiliary memory to enhance its performance \cite{wang2017recurrent}. \cite{chung2015gated} introduced a gated feedback RNN architecture which stacks multiple RNNs to pass/not pass and control signals flowing from upper layers to the lower layers. Another modification for RNNs is the Clockwork RNN \cite{KoutnikGGS14} which introduced a method that makes RNNs work for long-term sequences requirements. It was shown to outperform LSTMs and RNNs in  some specific tasks.\\

LSTMs have been originally developed by \cite{hochreiter1997long}. The main motivation of LSTMs is to skive the problem of vanishing gradients in vanilla RNNs and to remember longer sequences. The hidden layer of a RNN was replaced by a memory unit or what called a memory cell. The LSTM has specific function gates to control the flow of data and its storage within the memory cell. One extra gate added to the LSTM called the forget gate it has been introduced by \cite{gers1999learning} to give LSTM the ability to forget specific information from the memory cell. From this point, multiple developments has been done to improve the performance of LSTMs. One of them is to replace the feed forward units with a convolutional neural networks (CNNs) introduced by \cite{lecun1998gradient} to improve its ability for visual sequence problems \cite{donahue2015long}. Other approaches involved stacking LSTMs \cite{grave1,SutskeverVL14,graves2012supervised} Or by introducing a depth gate between stacked LSTMs \cite{yao2015depth}. Also, some proposed a hyper-architecture between RNNs and LSTMs such as the work of \cite{KrauseLMR16}. Nesting the LSTM within another LSTM, resulting in a nested LSTM (NLSTM) is the focus of \cite{NLSTM}, which is used as a reference in this article. \cite{KalchbrennerDG15} organized LSTM in the form of a multidimensional grid. An extensive work by \cite{greff2017lstm} explored different variations of LSTM by introducing six variants to the architecture. It concluded that the current LSTM is indeed performing well compared to these variants.  Also it found that the forget gate and output gate activation functions are very critical to the LSTM performance.\\

The GRUs architecture, which is on the par with LSTMs in terms of performance, was introduced by introduced by \cite{GRUSOURCE}. GRUs requires less memory and computationally less expensive than LSTM. GRUs in some cases may outperform LSTMs, as shown in the comparative study by \cite{ChungGCB14}. Also, GRUs fully expose the hidden state content unlike LSTMs. For the development of GRUs, an approach in the work done by \cite{shi2017learning} introduced the shuttleNet concept. The shuttleNet uses multiple GRUs treated as processors. These processors are loops connected to mimic the human’s brain feedback and feed-forward connections. A study on three different variants of GRUs was done by \cite{dey2017gate} concluded that the current GRUs have a similar performance as these three variants.\\

\section{Mother Compact Recurrent Memory (MCRM)} \label{MCRMSECTION}
In this section, we first define the notations used in this article. We then recall the LSTM and GRU models, and use them to derive the MCRM model.
\subsection{Mathematical notions}
The following mathematical notations will be used. \(t\) stands for current time step, \(\odot\) is the Hadamard product, \(\sigma\) is the sigmoid activation function and \(\tanh\) is the tanh activation function. \(\concat\) is the concatenation symbol defined in \cite{Wadler:1989:TF:99370.99404}. The input to any architecture is \(\mathbf x_t \in \mathbb{R}^{m}\). Weights which interacts with the input are \(W_{x\alpha} \in \mathbb{R}^{m\times p} \). Weights which interacts with the hidden state \(\mathbf h_t \in \mathbb{R}^{p}\) are \(W_{h\alpha} \in \mathbb{R}^{p\times p} \). \(b_{\alpha} \in \mathbb{R}^{p} \) are the biases. \( \alpha \in \{ \mathbf{i},\mathbf{f},\mathbf{c},\mathbf{o},\mathbf{h},\mathbf{r},\mathbf{z},\mathbf{n} \}\) represent the different gates in both LSTM and GRU equations \eqref{lstmeqn} ,\eqref{gruqen} respectively.

\subsection{LSTMs architecture}
LSTMs address the vanishing gradient problem commonly found in RNNs by controlling the information flow through specific functions gates \cite{hochreiter1997long}. At each time step, an LSTM maintains a hidden state vector $\mathbf h_t$ and a memory state vector $\mathbf c_t$  responsible for controlling the state updates and generating the outputs. The computation at time step $t$  is defined as follows:\\
\begin{equation}
  \begin{split} \label{lstmeqn}
    &\mathbf i_t =\sigma(\mathbf x_t*W_{xi} + \mathbf h_{t-1}*W_{hi} + b_i) \\
    &\mathbf f_t =\sigma(\mathbf x_t*W_{xf} + \mathbf h_{t-1}*W_{hf} + b_f) \\
    &\mathbf c_t =\mathbf f_t \odot \mathbf c_{t-1} +\mathbf i_t\odot \tanh(\mathbf x_t*W_{xc} + \mathbf h_{t-1}*W_{hc} + b_c) \\
    &\mathbf o_t =\sigma(\mathbf x_t*W_{xo} +\mathbf h_{t-1}*W_{ho} + b_o) \\
	&\mathbf h_t =o_t\odot\tanh(c_t)
\end{split}
\end{equation}
$\mathbf c_t$ is usually refereed as the cell or the memory state where the information are stored. The hidden state $\mathbf h_t$ is the output or the exposed state of an LSTM. The LSTM operation is as follows: the input gate \( \mathbf i_t \) at time step \(t\) decides how much information to take from the input \( \mathbf x_t\) into the memory. The forget gate \(\mathbf f_t\) decides how much information to keep from the previous cell state \( \mathbf c_{t-1} \). Both input gate interaction and forget gate interaction are used to compute the new cell state \( \mathbf c_t\). Then the output gate \(\mathbf o_t \) is used to compute the quantity of information to expose from the cell state \( \mathbf c_t\) to outside world through the hidden state \( \mathbf h_t\) representing the output of LSTM.\\

\subsection{GRUs architecture}
GRUs also address the same problems found in RNNs. They have been introduced by \cite{GRUSOURCE}. The main difference between GRUs and LSTMs is that GRUs totally expose the hidden state information through \(\mathbf h_t\). They require less memory and are computationally less expensive unlike LSTMs. The computation at time step $t$  is defined as follows:\\
\begin{equation}
  \begin{split} \label{gruqen}
  &\mathbf r_t = \sigma(W_{ir}*\mathbf x_t + b_{ir} + W_{hr}* \mathbf h_{t-1} + b_{hr}) \\
  &\mathbf z_t = \sigma(W_{iz}* \mathbf x_t + b_{iz} + W_{hz}* \mathbf h_{t-1} + b_{hz}) \\
  &\mathbf n_t = \tanh(W_{in} *\mathbf x_t + b_{in} + \mathbf r_t \odot(W_{hn}* \mathbf h_{t-1}+ b_{hn})) \\
  &\mathbf h_t = (1 - \mathbf z_t)\odot \mathbf h_{t-1} + \mathbf z_t \odot \mathbf n_t
\end{split}
\end{equation}
The GRU operates as follows: the reset gate \(\mathbf r_t\) is used to compute how much information to remove from the previous hidden state \(\mathbf h_{t-1}\). This reset gate interaction is added to the input \(\mathbf x_t\) and saved into an intermediate vessel called the node state \(\mathbf n_t\). The update gate \(\mathbf z_t\) decides how much information from the previous hidden state \(\mathbf h_{t-1}\) should be added to the node state \(\mathbf n_t\) to form the new hidden state \(\mathbf h_{t}\). This new hidden state \(\mathbf h_{t}\) is the output of the GRU cell.\\

\subsection{MCRMs architecture}
MCRM nests a GRU cell inside an LSTM treating the GRU hidden state \(\mathbf h_t^{GRU}\) as the LSTM cell state $\mathbf c_t$ .
The input to the GRU unit is \(\mathbf x_t^{GRU}\) which is the concatenation of what the LSTM should forget and what it should remember from the input coming into the MCRM cell, it's defined in equation \eqref{eq:eq_lstminput}. \\
\begin{equation}\label{eq:eq_lstminput}
  \begin{split}
    &\mathbf x_t^{GRU} =\concat(\mathbf f_t \odot \mathbf c_{t-1},\\
    &\mathbf i_t\odot \tanh(\mathbf x_t*W_{xc} + \mathbf h_{t-1}*W_{hc} + b_c)) 
\end{split}
\end{equation}

The modified equation of the LSTM cell now become: \\
\begin{equation}
  \begin{split}
    &\mathbf i_t =\sigma(\mathbf x_t*W_{xi} + \mathbf h_{t-1}*W_{hi} + b_i) \\
    &\mathbf f_t =\sigma(\mathbf x_t*W_{xf} + \mathbf h_{t-1}*W_{hf} + b_f) \\
    &\mathbf c_t =\mathbf h_t^{GRU} \\
    &\mathbf o_t =\sigma(\mathbf x_t*W_{xo} +\mathbf h_{t-1}*W_{ho} + b_o) \\
	&\mathbf h_t =o_t\odot\tanh(c_t)
\end{split} \label{MCRM_lstmeqn}
\end{equation} 

And the modified equations of GRU cell now becomes:\\
\begin{equation}
  \begin{split}
  &\mathbf r_t = \sigma(W_{ir}*\mathbf x_t^{GRU} + b_{ir} + W_{hr}* \mathbf h_{t-1}^{GRU} + b_{hr}) \\
  &\mathbf z_t = \sigma(W_{iz}* \mathbf x_t^{GRU} + b_{iz} + W_{hz}* \mathbf h_{t-1}^{GRU} + b_{hz}) \\
  &\mathbf n_t = \tanh(W_{in} *\mathbf x_t^{GRU} + b_{in} + \mathbf r_t \odot(W_{hn}* \mathbf h_{t-1}^{GRU}+ b_{hn})) \\
  &\mathbf h_t^{GRU} = (1 - \mathbf z_t)\odot \mathbf h_{t-1}^{GRU} + \mathbf z_t \odot \mathbf n_{t}
\end{split}\label{MCRM_grueqn}
\end{equation}

\begin{figure}[!t]
\begin{center}
\centerline{\includegraphics[width=\columnwidth]{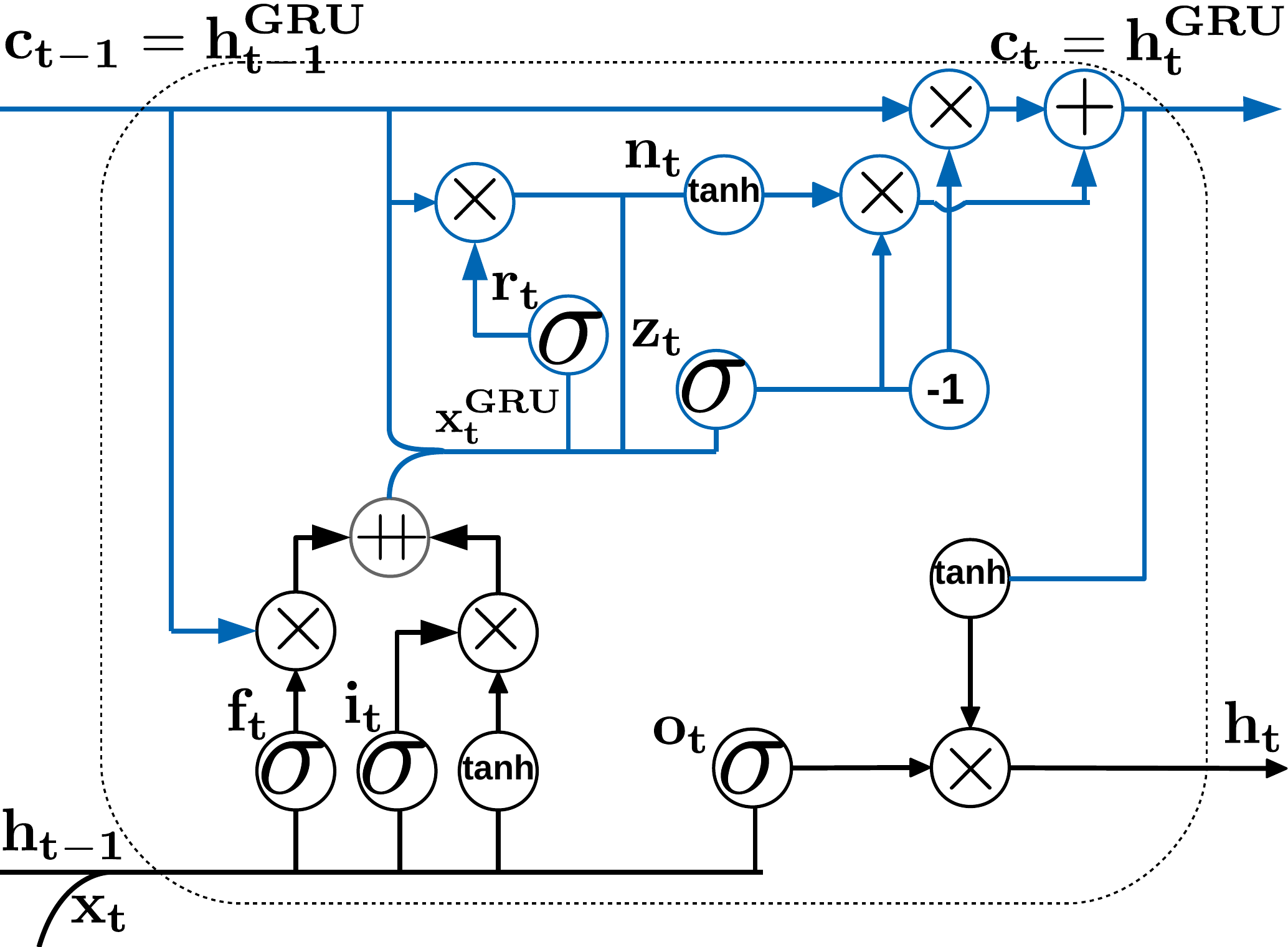}}
\caption{MCRM detailed data flow diagram starting from the previous hidden state to the next hidden state. The black colored lines refers to the LSTM part and the blue colored lines refers to the GRU part. }
\vskip 0.1in
\label{MCRM_data_flow}
\end{center}
\end{figure}

Figure \ref{MCRM_data_flow} illustrates the data flow inside the MCRM, following equations \ref{MCRM_lstmeqn} and \ref{MCRM_grueqn}. The closest architecture to MCRMs is the Nested LSTMs (NLSTMs) introduced by \cite{NLSTM}, which nests am LSTM inside another LSTM. MCRMs had the following advantages over NLSTMs: first, they are less computationally expensive than NLSTMs as they use GRUs instead of LSTM as the cell state. Secondly, they have a better neurons utilization as the full hidden state is exposed from the GRU to the LSTM unlike NLSTMs where the inner LSTM are not fully utilized because of the usage of the cell state only.\\ 

\section{Experiments and results } \label{experiments-sec}
In this section, the MCRM performance is evaluated empirically against different recurrent architectures on well-known tasks. For fairness of comparison, we use the same hyper-parameters as in \cite{bai2018empirical}. Also, our results are consist with the results from \cite{bai2018empirical}. These experiments were executed under a controlled environment, using the same initial random seed and weights initializations. The parameters of the models were kept the same between multiple experiments to check which architecture has a better usage of the neurons. The different configuration parameters are shown in table \ref{TBL_SETTINGS}. Each experiment was executed multiple times with different initial seeds and the reported performance metrics are the mean performance of these multiple executions. We avoided the usage of any drop-out or batch normalization layers to have a fair evaluation of the performance.\\

\begin{table*}[t]
\def\arraystretch{1.3}
\small
\centering
\caption{Evaluation of MCRM versus different recurrent architectures on synthetic stress tests, character-level language modeling, and word-level language modeling. The MCRM architecture outperforms most of these recurrent networks in some cases or tend to be better than other architectures across different tasks and datasets. $\ {}^h$~means that higher is better. ${}^\ell$~means that lower is better.}
\vspace{2mm}
{\begin{tabular}{l*{6}{c}}
\toprule
\multirow{2}{*}{Sequence Modeling Task} & \multirow{2}{*}{Model Size ($\approx$)} & \multicolumn{5}{c}{Models} \\
\cline{3-7}
\textbf    & & RNN & GRU & LSTM & NSLTM & \textbf{MCRM} \\
\midrule
Seq. MNIST (accuracy${}^h$)             & 152K  & 19.57   & 98.58   & 85.16   & 91.02   &\textbf{98.79}   \\
Adding problem (loss${}^\ell$)          & 95K   & 0.165 & 3.2e-04 & 0.001   & 0.004   & \textbf{4.0e-06} \\
Copy memory (loss)                      & 3.3M  &0.021     & 0.013  & 0.004 & 7.3e-05 & \textbf{8.5e-06}\\
Char-level PTB (bpc${}^\ell$)      		& 17.1M    & 1.683     & 1.397 & 1.374     & 1.365    &  \textbf{1.331}  \\
Word-level PTB (ppl${}^\ell$)    		& 1.3M    & 140.58 & \textbf{110.6}  & 110.64 & 140.1  & 120.9 \\
\bottomrule
\end{tabular}}
\label{TBL_RESULTS}
\end{table*}

\begin{figure}[t!]
\begin{center}
\centerline{\includegraphics[width=\columnwidth]{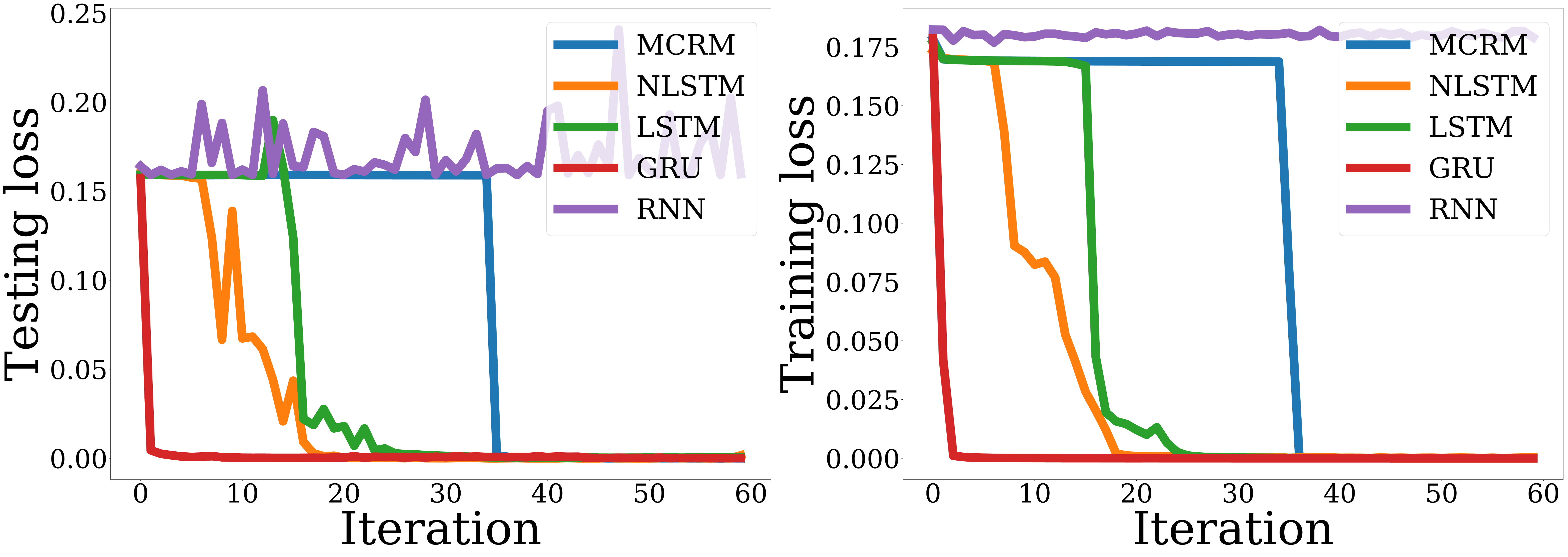}}
\caption{The loss as a function of iteration for the adding problem test predictions for both  train and test datasets.}
\label{FIG_ADDING_200}
\end{center}
\end{figure}

\paragraph{The adding problem}
The adding problem has been used as a stress test for sequence models. It was introduced by \cite{hochreiter1997long}. The test is about creating an input with a length \(T\) sequence of depth 2. The first dimension is randomly chosen between 0 and 1. The second dimension is all zeros except the last two elements marked by 1. The objective is find the sum the last two random elements marked by 1 in the second dimension. We used a sequence of length \(T = 200\). The test results are shown in table \ref{TBL_RESULTS}. MCRM outperforms all other models with an error of \(4e-06\). Also, the GRU has a close performance of an error of \(3.2e-04\) which explains why the MCRM have this performance. The learning curves are shown in figure \ref{FIG_ADDING_200}.

\begin{figure}[t!]
\begin{center}
\centerline{\includegraphics[width=\columnwidth]{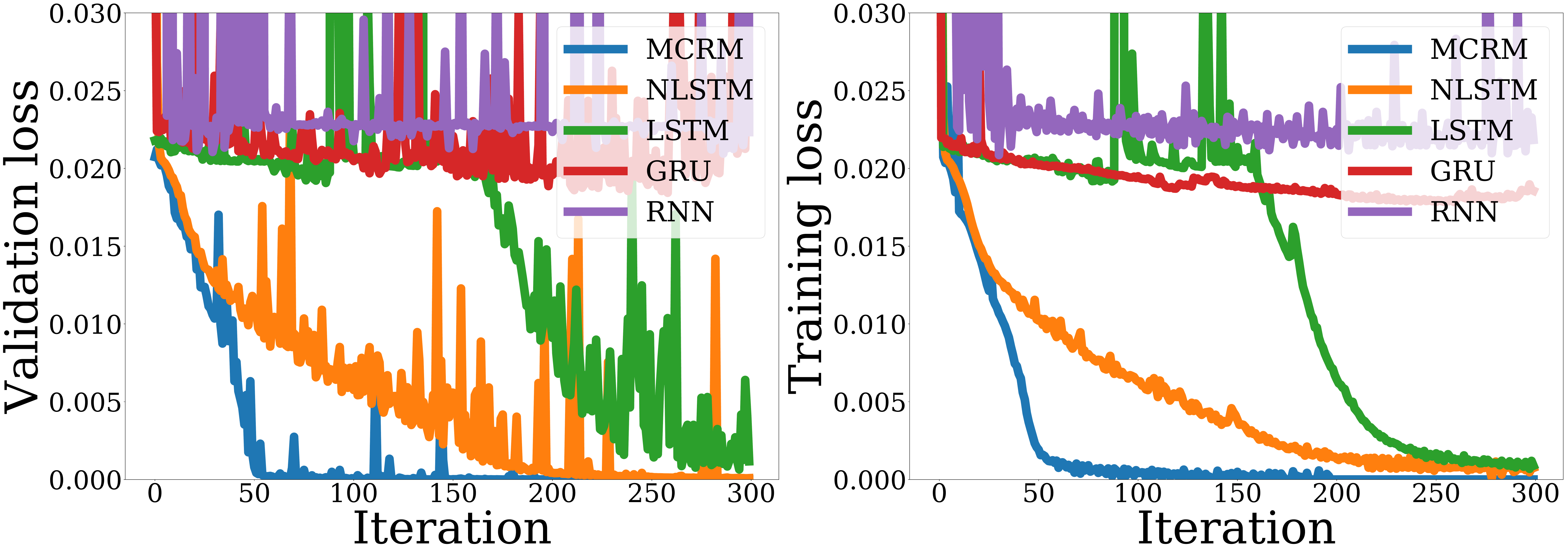}}
\caption{The loss as a function of iteration for the copy memory test predictions for both  train and test datasets.}
\label{copymem_1}
\end{center}
\end{figure}

\paragraph{Copy memory test}
This method has been used previously in \cite{ZhangWCLMSB16,ArjovskySB15,JingSDPSTS16} for measuring the performance of a recurrent architecture in the context of remembering information seen \(T\) time steps earlier. The input sequence is in the length of \(T+20\). The input sequence defined as: \( \{ \beta_0,\dots,\beta_9, \kappa_0,\dots,\kappa_n,\delta_{\text{delm}},\delta_0,\dots,\delta_9 \} \), where \(\kappa \) is chosen to be a zero digit and \(n =T-1 \). The \(\beta\) is randomly chosen from digits \(\{1,\dots,8 \} \). The \( \delta \) is set to be digit 9, where \(\delta_{\text{delm}}\) is the delimiter. The model is expected to generate an output identical to the input sequence is. A test was conducted with a sequence length of \(T=1000\). The results are shown in table \ref{TBL_RESULTS}. The MCRM outperforms all other models with an error of \(8.5e-06\). The learning curves are shown in figure \ref{copymem_1}.

\begin{figure}[t!]
\begin{center}
\centerline{\includegraphics[width=\columnwidth]{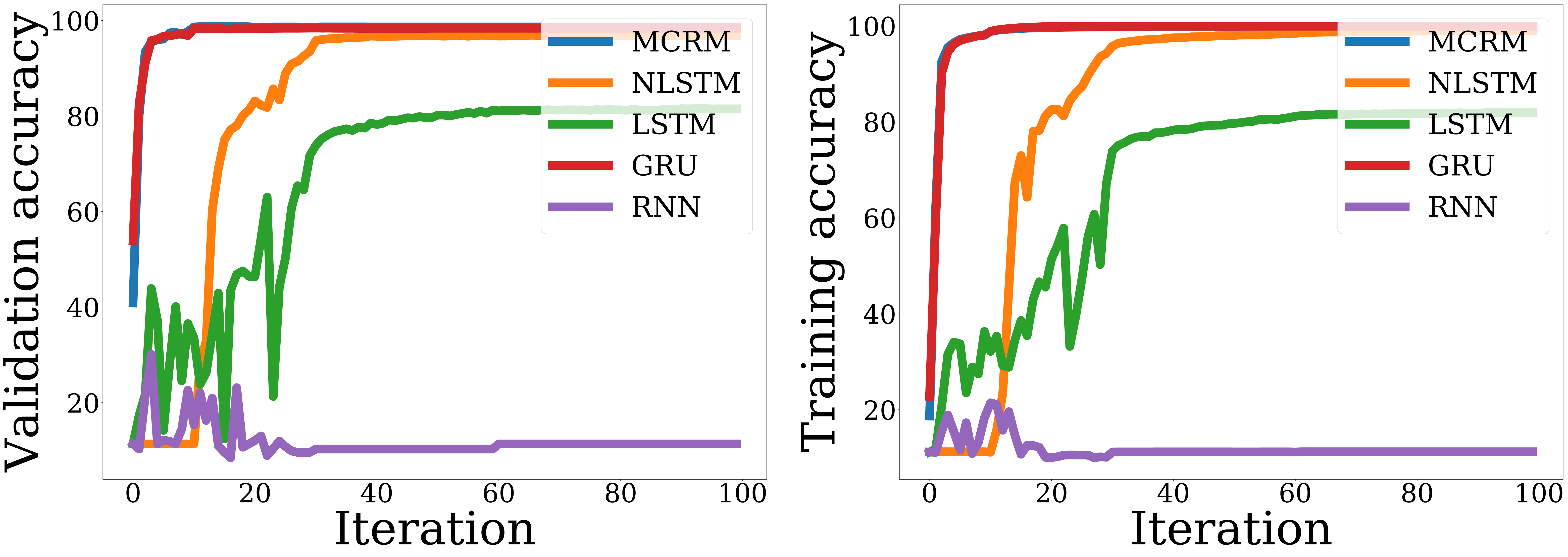}}
\caption{The accuracy as a function of iteration for the sequential MNIST predictions for both  train and
test datasets.}
\label{smnist_2}
\end{center}
\end{figure}

\paragraph{Sequential MNIST test} 
This test is similar in intent to the copy memory test. In this task the MNIST dataset \cite{lecun2010mnist} images are presented to the model as a \(784\times 1\) input sequence of pixels intensity values. The recurrent model should be able to reconstruct the image again. This test was used as a stress test in several recurrent related problems \cite{le2015simple,zhang2016architectural}. From table \ref{TBL_RESULTS} MCRM achieved accuracy of 98.79\% outperforming any other model. Surprisingly, the LSTM and NSLTM had a poor performance unlike the GRU. We relate the success of MCRM in this task to the GRU core inside it. The learning curves are presented in figure \ref{smnist_2}.

\paragraph{PennTreebank character and word levels tests} 

\begin{figure}[t!]
\begin{center}
\centerline{\includegraphics[width=\columnwidth]{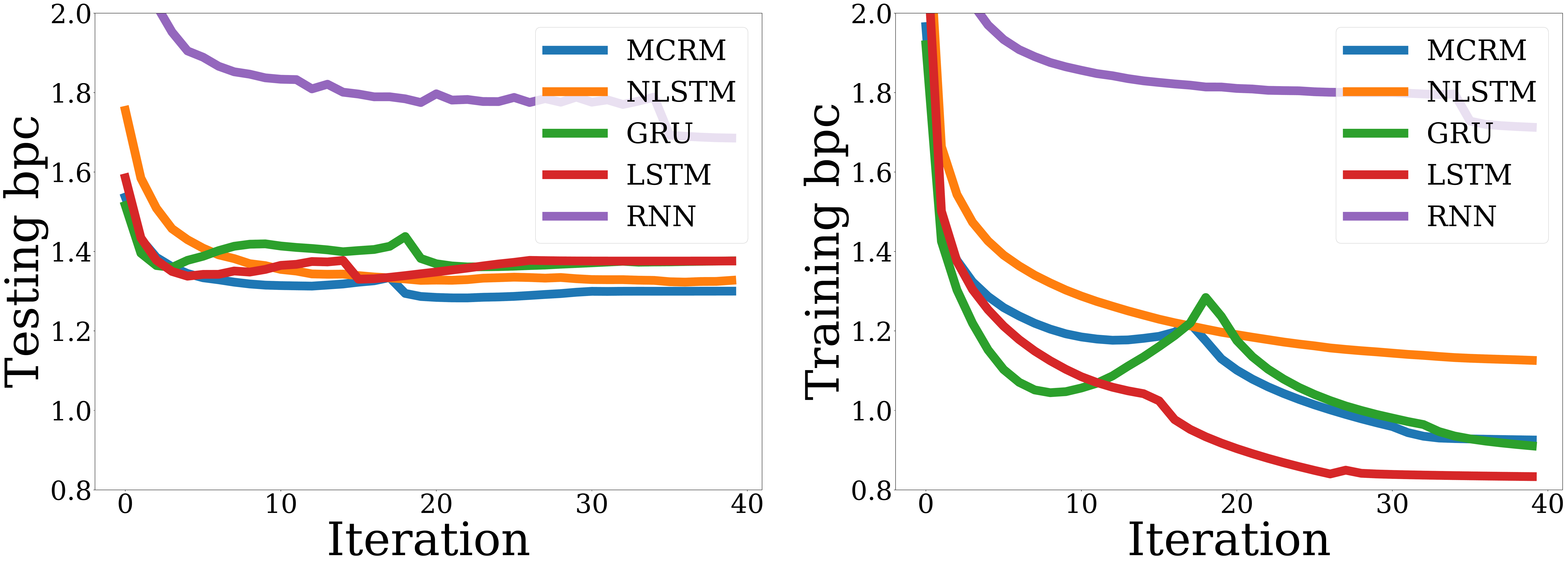}}
\caption{The bpc performance index as a function of iteration for character-level prediction on PTB’s train and
test data sets.}
\label{char_ptb_2}
\end{center}
\end{figure}

\begin{figure}[t!]
\begin{center}
\centerline{\includegraphics[width=\columnwidth]{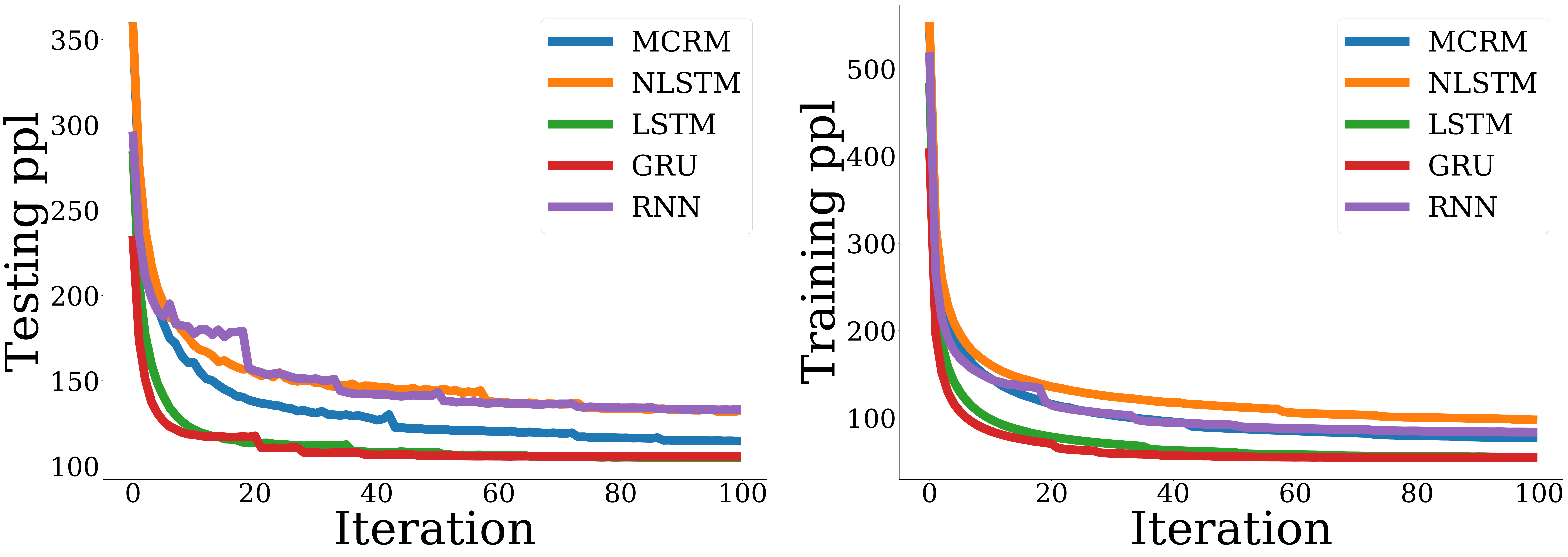}}
\caption{The ppl performance index as a function of iteration for word-level prediction on PTB’s train and
test data sets.}
\label{word_ptb_2}
\end{center}
\end{figure}

The PennTreebank (PTB) \cite{marcus1993building} is a text data set for both character-level and word-level language modeling tasks. It is widely used in many RNN architectures for evaluating the model performance. The PTB is divided into train, test and validation datasets. To measure the character-level task performance the bits per character (bpc) is used as a performance index. BPC has been introduced by \cite{Graves13}and it is defined as the cross-entropy loss \cite{Goodfellow-et-al-2016} divided by \(log \text{ 2} \). The performance index for the word-level language modeling task is the perplexity (ppl). The ppl is defined as the exponential of the cross-entropy loss. The two tasks results are reported in table \ref{TBL_RESULTS}. The reported values are from the validation dataset.
When the PTB is used as character-level language corpus the MCRM outperforms other models with a bpc of 1.331 exceeding the NLSTM by 0.034 ppl. The learning curves are shown in figure \ref{char_ptb_2}. When the PTB used as word-level language corpus MCRM performance is in-between the GRU and NLSTM. The learning curves are shown in figure \ref{word_ptb_2}

\begin{table*} \label{exp-param-table}
\def\arraystretch{1.3}
\small
\centering
\caption{Parameter settings for experiments in Section \ref{experiments-sec}.
The integers beneath each architecture type are the hidden state sizes. \({lr}\) stands for learning rate. \({gc}\) stands for gradient clipping}
{\begin{tabular}{ |c|c|c|c|c|c|c|c| }
\hline
\multicolumn{8}{|c|}{\textsc{Key Parameters}} \\
\hline
\textbf{Dataset/Task} & $\textbf{RNN}$ & \textbf{GRU} & \textbf{LSTM} & \textbf{NLSTM} & \textbf{MCRM} & \textbf{\({lr}\), \({gc}\)} & \textbf{Note}   \\ \hline
\multirow{1}{*}{Adding}  & 308 & 177 & 153 & 77 & 85 & Adam \({lr}\)(1e-3), \({gc}\)(0.5) &NLSTM \({lr}\)(0.01), \({gc}\)(0.1) \\
\hline
Seq. MNIST    						 & 384 &222 & 192 & 108 & 97 & RMSprop \({lr}\)(1e-3), \({gc}\)(1.0) &NLSTM \({gc}\)(0.25), LSTM \({lr}\)(1e-4)  \\
\hline
\multirow{1}{*}{Copy Memory}    &  1800 & 1050 & 900 & 448 & 500 & RMSprop \({lr}\)(5e-4), \({gc}\)(1.0)  &NLSTM \({lr}\)(1e-4), \({gc}\)(0.25)\\

\hline
\multirow{1}{*}{Word-level PTB} 	 &  125 & 119 & 117 & 100 & 109 & SGD \({lr}\)(30), \({gc}\)(0.35) & -\\

\hline
\multirow{1}{*}{Char-level PTB} 	& 2900 & 1680 & 1050 & 920 & 1000 &  Adam \({lr}\)(1e-3), \({gc}\)(0.15) & - \\
\hline
\end{tabular}}
\label{TBL_SETTINGS}
\end{table*}

\section{Visualization} \label{visualizationsec}
\begin{figure*}
\includegraphics[width=\textwidth]{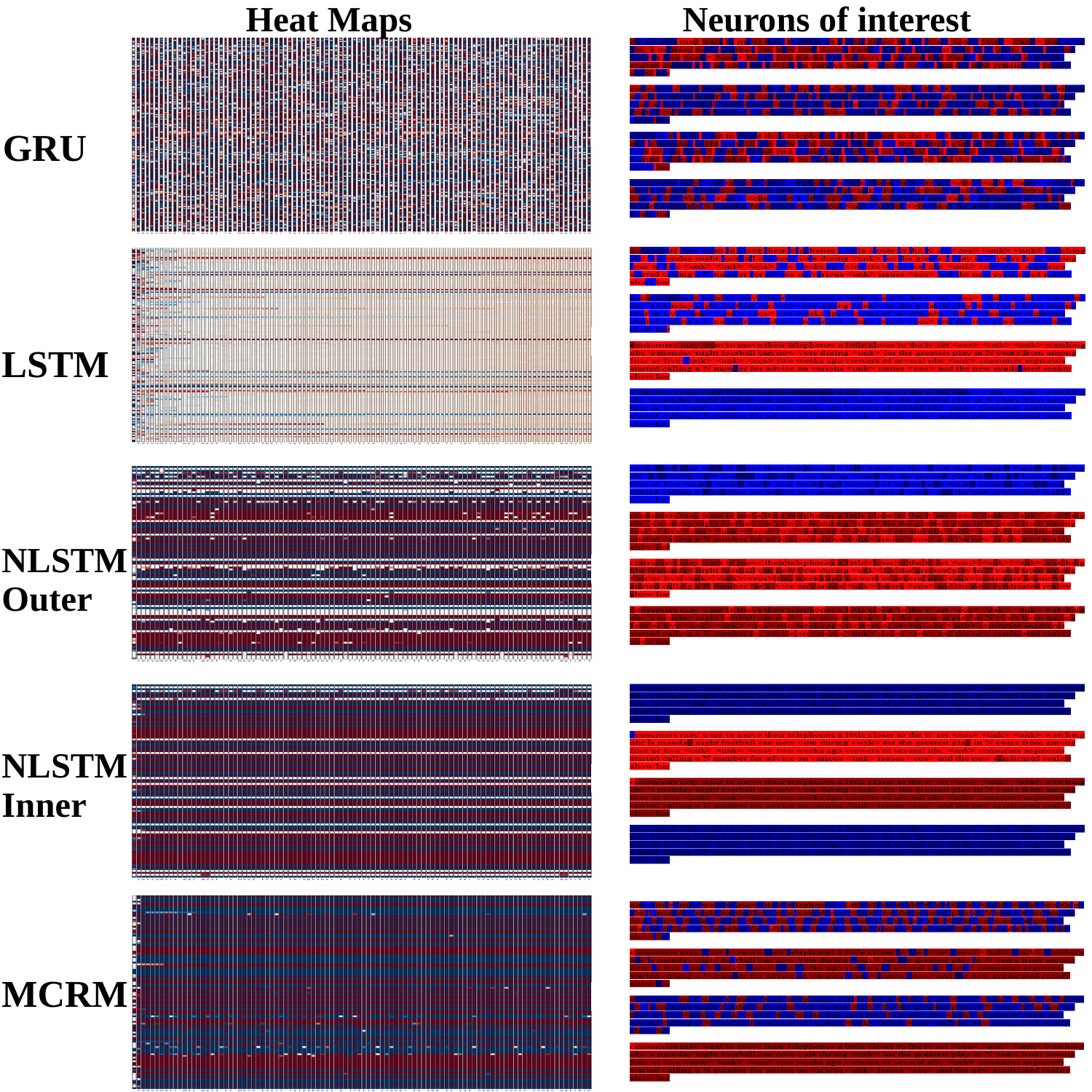}
\caption{Heat Maps column is the heat maps of the propagation of all memory states neurons versus sequence of character inputs. The vertical axis represents neurons activation values versus the horizontal axis. The horizontal axis represents the characters input to the cell. Each row is the propagation of a specific neuron activation values.The propagation is from left to right. The neurons of interest columns is a visualization of specific neurons of interest activation values versus an input sequence of characters. Red denotes a negative cell state value, and blue a positive one. A darker shade denotes a larger magnitude. The memory state of GRU  is its hidden cell state \(h_t\). The memory state of LSTM is its cell state \(C_t\). The memory state of NLSTM are its outer \(C_t^{outer}\) and inner \(tanh(C_t^{inner})\) cell states respectively. The memory state of MCRM cell is its cell state \(h_t^{GRU}\).}
\label{fig_mem_text}
\end{figure*}

To understand the internal behavior of MCRM, we performed a visual analysis of the memory cell. Following a similar approach as the work of \cite{KarpathyJL15}, specific neurons of interest are monitored versus an input sequence. The work of \cite{KarpathyJL15} is also expanded by introducing a method of selecting these neurons. This method consists of a heat map of the propagation of all neurons activation values in the memory in contrast to an input sequence (shown in figure \ref{fig_mem_text}.). MCRM, NLSTM, LSTM and GRU are trained over the PTB character dataset, fixing the memory cell size to be around 150 neurons.\\

In figure \ref{fig_mem_text} the heat maps columns represents a step in this visual analysis. LSTM cell states tend to have neurons changing slowly over the sequence. This is in contrast to GRU architectures where-in neuron activation values change rapidly. NLSTM outer cell acts in a short-term fashion remembering small sequences. The inner cell of the NLSTM acts in a long-term fashion to support longer sequences. MCRM memory cell has some neurons acting in an explicit long-term fashion and some acting in a short-term explicit fashion. This means by nature MCRM inherits both LSTM and GRU behaviors in a one memory cell. This leads to a better neurons utilization which is an important advantage of MCRM. The neurons of interest column in Figure \ref{fig_mem_text} shows specific neurons extracted from the heat maps column that have long or short-term behaviors support the analysis of the heat maps. 

\section{Conclusion}

Mother Compact Recurrent Memory (MCRMs) are a Nested LSTM-GRU architecture. They create a unique compact memory pattern that supports both long and short-terms behaviors. MCRMs can outperform other RNNs architectures, on benchmark tests. Because of their promising results, MCRMs could be used in temporal sequence modeling tasks.

\section*{Acknowledgements}
This work was partially supported by the National Science Foundation under grant 1739964: CPS: Medium: Augmented reality for control of reservation-based intersections with mixed autonomous-non autonomous flows.

\bibliography{MCRM_main}

\begin{thebibliography}{}

\bibitem[\protect\citeauthoryear{Arik \bgroup et al\mbox.\egroup
  }{2017}]{arik2017deep}
Arik, S.~O.; Chrzanowski, M.; Coates, A.; Diamos, G.; Gibiansky, A.; Kang, Y.;
  Li, X.; Miller, J.; Ng, A.; Raiman, J.; et~al.
\newblock 2017.
\newblock Deep voice: Real-time neural text-to-speech.
\newblock {\em arXiv preprint arXiv:1702.07825}.

\bibitem[\protect\citeauthoryear{Arjovsky, Shah, and
  Bengio}{2015}]{ArjovskySB15}
Arjovsky, M.; Shah, A.; and Bengio, Y.
\newblock 2015.
\newblock Unitary evolution recurrent neural networks.
\newblock {\em CoRR} abs/1511.06464.

\bibitem[\protect\citeauthoryear{Bahdanau, Cho, and
  Bengio}{2014}]{bahdanau2014neural}
Bahdanau, D.; Cho, K.; and Bengio, Y.
\newblock 2014.
\newblock Neural machine translation by jointly learning to align and
  translate.
\newblock {\em arXiv preprint arXiv:1409.0473}.

\bibitem[\protect\citeauthoryear{Bai, Kolter, and
  Koltun}{2018}]{bai2018empirical}
Bai, S.; Kolter, J.~Z.; and Koltun, V.
\newblock 2018.
\newblock An empirical evaluation of generic convolutional and recurrent
  networks for sequence modeling.
\newblock {\em arXiv preprint arXiv:1803.01271}.

\bibitem[\protect\citeauthoryear{Cho \bgroup et al\mbox.\egroup
  }{2014}]{GRUSOURCE}
Cho, K.; van Merrienboer, B.; G{\"{u}}l{\c{c}}ehre, {\c{C}}.; Bougares, F.;
  Schwenk, H.; and Bengio, Y.
\newblock 2014.
\newblock Learning phrase representations using {RNN} encoder-decoder for
  statistical machine translation.
\newblock {\em CoRR} abs/1406.1078.

\bibitem[\protect\citeauthoryear{Chung \bgroup et al\mbox.\egroup
  }{2014}]{ChungGCB14}
Chung, J.; G{\"{u}}l{\c{c}}ehre, {\c{C}}.; Cho, K.; and Bengio, Y.
\newblock 2014.
\newblock Empirical evaluation of gated recurrent neural networks on sequence
  modeling.
\newblock {\em CoRR} abs/1412.3555.

\bibitem[\protect\citeauthoryear{Chung \bgroup et al\mbox.\egroup
  }{2015}]{chung2015gated}
Chung, J.; Gulcehre, C.; Cho, K.; and Bengio, Y.
\newblock 2015.
\newblock Gated feedback recurrent neural networks.
\newblock In {\em International Conference on Machine Learning},  2067--2075.

\bibitem[\protect\citeauthoryear{Dey and Salem}{2017}]{dey2017gate}
Dey, R., and Salem, F.~M.
\newblock 2017.
\newblock Gate-variants of gated recurrent unit (gru) neural networks.
\newblock {\em arXiv preprint arXiv}.

\bibitem[\protect\citeauthoryear{Donahue \bgroup et al\mbox.\egroup
  }{2015}]{donahue2015long}
Donahue, J.; Anne~Hendricks, L.; Guadarrama, S.; Rohrbach, M.; Venugopalan, S.;
  Saenko, K.; and Darrell, T.
\newblock 2015.
\newblock Long-term recurrent convolutional networks for visual recognition and
  description.
\newblock In {\em Proceedings of the IEEE conference on computer vision and
  pattern recognition},  2625--2634.

\bibitem[\protect\citeauthoryear{Elman}{1990}]{elman1990finding}
Elman, J.~L.
\newblock 1990.
\newblock Finding structure in time.
\newblock {\em Cognitive science} 14(2):179--211.

\bibitem[\protect\citeauthoryear{Fan \bgroup et al\mbox.\egroup
  }{2014}]{fan2014tts}
Fan, Y.; Qian, Y.; Xie, F.-L.; and Soong, F.~K.
\newblock 2014.
\newblock Tts synthesis with bidirectional lstm based recurrent neural
  networks.
\newblock In {\em Fifteenth Annual Conference of the International Speech
  Communication Association}.

\bibitem[\protect\citeauthoryear{Gers, Schmidhuber, and
  Cummins}{1999}]{gers1999learning}
Gers, F.~A.; Schmidhuber, J.; and Cummins, F.
\newblock 1999.
\newblock Learning to forget: Continual prediction with lstm.

\bibitem[\protect\citeauthoryear{Goodfellow, Bengio, and
  Courville}{2016}]{Goodfellow-et-al-2016}
Goodfellow, I.; Bengio, Y.; and Courville, A.
\newblock 2016.
\newblock {\em Deep Learning}.
\newblock MIT Press.
\newblock \url{http://www.deeplearningbook.org}.

\bibitem[\protect\citeauthoryear{Graves, Mohamed, and Hinton}{2013}]{grave1}
Graves, A.; Mohamed, A.; and Hinton, G.~E.
\newblock 2013.
\newblock Speech recognition with deep recurrent neural networks.
\newblock {\em CoRR} abs/1303.5778.

\bibitem[\protect\citeauthoryear{Graves}{2012}]{graves2012supervised}
Graves, A.
\newblock 2012.
\newblock Supervised sequence labelling.
\newblock In {\em Supervised sequence labelling with recurrent neural
  networks}. Springer.
\newblock  5--13.

\bibitem[\protect\citeauthoryear{Graves}{2013}]{Graves13}
Graves, A.
\newblock 2013.
\newblock Generating sequences with recurrent neural networks.
\newblock {\em CoRR} abs/1308.0850.

\bibitem[\protect\citeauthoryear{Greff \bgroup et al\mbox.\egroup
  }{2017}]{greff2017lstm}
Greff, K.; Srivastava, R.~K.; Koutn{\'\i}k, J.; Steunebrink, B.~R.; and
  Schmidhuber, J.
\newblock 2017.
\newblock Lstm: A search space odyssey.
\newblock {\em IEEE transactions on neural networks and learning systems}
  28(10):2222--2232.

\bibitem[\protect\citeauthoryear{Hochreiter and
  Schmidhuber}{1997}]{hochreiter1997long}
Hochreiter, S., and Schmidhuber, J.
\newblock 1997.
\newblock Long short-term memory.
\newblock {\em Neural computation} 9(8):1735--1780.

\bibitem[\protect\citeauthoryear{Jing \bgroup et al\mbox.\egroup
  }{2016}]{JingSDPSTS16}
Jing, L.; Shen, Y.; Dubcek, T.; Peurifoy, J.; Skirlo, S.~A.; Tegmark, M.; and
  Soljacic, M.
\newblock 2016.
\newblock Tunable efficient unitary neural networks {(EUNN)} and their
  application to {RNN}.
\newblock {\em CoRR} abs/1612.05231.

\bibitem[\protect\citeauthoryear{Jordan}{1997}]{jordan1997serial}
Jordan, M.~I.
\newblock 1997.
\newblock Serial order: A parallel distributed processing approach.
\newblock In {\em Advances in psychology}, volume 121. Elsevier.
\newblock  471--495.

\bibitem[\protect\citeauthoryear{Kalchbrenner, Danihelka, and
  Graves}{2015}]{KalchbrennerDG15}
Kalchbrenner, N.; Danihelka, I.; and Graves, A.
\newblock 2015.
\newblock Grid long short-term memory.
\newblock {\em CoRR} abs/1507.01526.

\bibitem[\protect\citeauthoryear{Kang \bgroup et al\mbox.\egroup
  }{2017}]{kang2017object}
Kang, K.; Li, H.; Xiao, T.; Ouyang, W.; Yan, J.; Liu, X.; and Wang, X.
\newblock 2017.
\newblock Object detection in videos with tubelet proposal networks.
\newblock In {\em Proc. CVPR}, volume~2, ~7.

\bibitem[\protect\citeauthoryear{Karpathy and Fei-Fei}{2015}]{karpathy2015deep}
Karpathy, A., and Fei-Fei, L.
\newblock 2015.
\newblock Deep visual-semantic alignments for generating image descriptions.
\newblock In {\em Proceedings of the IEEE conference on computer vision and
  pattern recognition},  3128--3137.

\bibitem[\protect\citeauthoryear{Karpathy, Johnson, and
  Li}{2015}]{KarpathyJL15}
Karpathy, A.; Johnson, J.; and Li, F.
\newblock 2015.
\newblock Visualizing and understanding recurrent networks.
\newblock {\em CoRR} abs/1506.02078.

\bibitem[\protect\citeauthoryear{Kong \bgroup et al\mbox.\egroup
  }{2018}]{AAAI1817074}
Kong, Y.; Gao, S.; Sun, B.; and Fu, Y.
\newblock 2018.
\newblock Action prediction from videos via memorizing hard-to-predict samples.

\bibitem[\protect\citeauthoryear{Koutn{\'{\i}}k \bgroup et al\mbox.\egroup
  }{2014}]{KoutnikGGS14}
Koutn{\'{\i}}k, J.; Greff, K.; Gomez, F.~J.; and Schmidhuber, J.
\newblock 2014.
\newblock A clockwork {RNN}.
\newblock {\em CoRR} abs/1402.3511.

\bibitem[\protect\citeauthoryear{Krause \bgroup et al\mbox.\egroup
  }{2016}]{KrauseLMR16}
Krause, B.; Lu, L.; Murray, I.; and Renals, S.
\newblock 2016.
\newblock Multiplicative {LSTM} for sequence modelling.
\newblock {\em CoRR} abs/1609.07959.

\bibitem[\protect\citeauthoryear{Le, Jaitly, and Hinton}{2015}]{le2015simple}
Le, Q.~V.; Jaitly, N.; and Hinton, G.~E.
\newblock 2015.
\newblock A simple way to initialize recurrent networks of rectified linear
  units.
\newblock {\em arXiv preprint arXiv:1504.00941}.

\bibitem[\protect\citeauthoryear{LeCun \bgroup et al\mbox.\egroup
  }{1998}]{lecun1998gradient}
LeCun, Y.; Bottou, L.; Bengio, Y.; and Haffner, P.
\newblock 1998.
\newblock Gradient-based learning applied to document recognition.
\newblock {\em Proceedings of the IEEE} 86(11):2278--2324.

\bibitem[\protect\citeauthoryear{LeCun, Cortes, and
  Burges}{2010}]{lecun2010mnist}
LeCun, Y.; Cortes, C.; and Burges, C.
\newblock 2010.
\newblock Mnist handwritten digit database.
\newblock {\em AT\&T Labs [Online]. Available: http://yann. lecun.
  com/exdb/mnist} 2.

\bibitem[\protect\citeauthoryear{Li \bgroup et al\mbox.\egroup
  }{2017}]{AAAI1714880}
Li, L.; Tang, S.; Deng, L.; Zhang, Y.; and Tian, Q.
\newblock 2017.
\newblock Image caption with global-local attention.

\bibitem[\protect\citeauthoryear{Lipton}{2015}]{Lipton15}
Lipton, Z.~C.
\newblock 2015.
\newblock A critical review of recurrent neural networks for sequence learning.
\newblock {\em CoRR} abs/1506.00019.

\bibitem[\protect\citeauthoryear{Lu, Lu, and Tang}{2017}]{lu2017online}
Lu, Y.; Lu, C.; and Tang, C.-K.
\newblock 2017.
\newblock Online video object detection using association lstm.
\newblock In {\em Proceedings of the IEEE Conference on Computer Vision and
  Pattern Recognition},  2344--2352.

\bibitem[\protect\citeauthoryear{Luong and Manning}{2015}]{luong2015stanford}
Luong, M.-T., and Manning, C.~D.
\newblock 2015.
\newblock Stanford neural machine translation systems for spoken language
  domains.
\newblock In {\em Proceedings of the International Workshop on Spoken Language
  Translation},  76--79.

\bibitem[\protect\citeauthoryear{Luong, Pham, and
  Manning}{2015}]{luong2015effective}
Luong, M.-T.; Pham, H.; and Manning, C.~D.
\newblock 2015.
\newblock Effective approaches to attention-based neural machine translation.
\newblock {\em arXiv preprint arXiv:1508.04025}.

\bibitem[\protect\citeauthoryear{Marcus, Marcinkiewicz, and
  Santorini}{1993}]{marcus1993building}
Marcus, M.~P.; Marcinkiewicz, M.~A.; and Santorini, B.
\newblock 1993.
\newblock Building a large annotated corpus of english: The penn treebank.
\newblock {\em Computational linguistics} 19(2):313--330.

\bibitem[\protect\citeauthoryear{Moniz and Krueger}{2018}]{NLSTM}
Moniz, J. R.~A., and Krueger, D.
\newblock 2018.
\newblock Nested lstms.
\newblock {\em CoRR} abs/1801.10308.

\bibitem[\protect\citeauthoryear{Niu \bgroup et al\mbox.\egroup
  }{2017}]{niu2017hierarchical}
Niu, Z.; Zhou, M.; Wang, L.; Gao, X.; and Hua, G.
\newblock 2017.
\newblock Hierarchical multimodal lstm for dense visual-semantic embedding.
\newblock In {\em Proceedings of the IEEE Conference on Computer Vision and
  Pattern Recognition},  1881--1889.

\bibitem[\protect\citeauthoryear{Rao \bgroup et al\mbox.\egroup
  }{2015}]{rao2015grapheme}
Rao, K.; Peng, F.; Sak, H.; and Beaufays, F.
\newblock 2015.
\newblock Grapheme-to-phoneme conversion using long short-term memory recurrent
  neural networks.
\newblock In {\em Acoustics, Speech and Signal Processing (ICASSP), 2015 IEEE
  International Conference on},  4225--4229.
\newblock IEEE.

\bibitem[\protect\citeauthoryear{Shi \bgroup et al\mbox.\egroup
  }{2017}]{shi2017learning}
Shi, Y.; Tian, Y.; Wang, Y.; Zeng, W.; and Huang, T.
\newblock 2017.
\newblock Learning long-term dependencies for action recognition with a
  biologically-inspired deep network.
\newblock In {\em Proceedings of the IEEE Conference on Computer Vision and
  Pattern Recognition},  716--725.

\bibitem[\protect\citeauthoryear{Sutskever, Vinyals, and
  Le}{2014}]{SutskeverVL14}
Sutskever, I.; Vinyals, O.; and Le, Q.~V.
\newblock 2014.
\newblock Sequence to sequence learning with neural networks.
\newblock {\em CoRR} abs/1409.3215.

\bibitem[\protect\citeauthoryear{Tripathi \bgroup et al\mbox.\egroup
  }{2016}]{tripathi2016context}
Tripathi, S.; Lipton, Z.~C.; Belongie, S.; and Nguyen, T.
\newblock 2016.
\newblock Context matters: Refining object detection in video with recurrent
  neural networks.
\newblock {\em arXiv preprint arXiv:1607.04648}.

\bibitem[\protect\citeauthoryear{Vinyals \bgroup et al\mbox.\egroup
  }{2015}]{vinyals2015show}
Vinyals, O.; Toshev, A.; Bengio, S.; and Erhan, D.
\newblock 2015.
\newblock Show and tell: A neural image caption generator.
\newblock In {\em Computer Vision and Pattern Recognition (CVPR), 2015 IEEE
  Conference on},  3156--3164.
\newblock IEEE.

\bibitem[\protect\citeauthoryear{Wadler}{1989}]{Wadler:1989:TF:99370.99404}
Wadler, P.
\newblock 1989.
\newblock Theorems for free!
\newblock In {\em Proceedings of the Fourth International Conference on
  Functional Programming Languages and Computer Architecture}, FPCA '89,
  347--359.
\newblock New York, NY, USA: ACM.

\bibitem[\protect\citeauthoryear{Wang \bgroup et al\mbox.\egroup
  }{2017}]{wang2017recurrent}
Wang, J.; Zhang, L.; Guo, Q.; and Yi, Z.
\newblock 2017.
\newblock Recurrent neural networks with auxiliary memory units.
\newblock {\em IEEE transactions on neural networks and learning systems}.

\bibitem[\protect\citeauthoryear{Wolf}{2017}]{wolf2017recurrent}
Wolf, C.
\newblock 2017.
\newblock Recurrent neural networks for object detection and motion
  recognition.

\bibitem[\protect\citeauthoryear{Xiong \bgroup et al\mbox.\egroup
  }{2018}]{AAAI1816788}
Xiong, H.; He, Z.; Hu, X.; and Wu, H.
\newblock 2018.
\newblock Multi-channel encoder for neural machine translation.

\bibitem[\protect\citeauthoryear{Yang \bgroup et al\mbox.\egroup
  }{2017}]{AAAI1714151}
Yang, M.; Tu, W.; Wang, J.; Xu, F.; and Chen, X.
\newblock 2017.
\newblock Attention based lstm for target dependent sentiment classification.

\bibitem[\protect\citeauthoryear{Yao \bgroup et al\mbox.\egroup
  }{2015}]{yao2015depth}
Yao, K.; Cohn, T.; Vylomova, K.; Duh, K.; and Dyer, C.
\newblock 2015.
\newblock Depth-gated recurrent neural networks.
\newblock {\em arXiv preprint}.

\bibitem[\protect\citeauthoryear{Yuan \bgroup et al\mbox.\egroup
  }{2017}]{yuan2017temporal}
Yuan, Y.; Liang, X.; Wang, X.; Yeung, D.~Y.; and Gupta, A.
\newblock 2017.
\newblock Temporal dynamic graph lstm for action-driven video object detection.
\newblock {\em arXiv preprint arXiv:1708.00666}.

\bibitem[\protect\citeauthoryear{Zhang \bgroup et al\mbox.\egroup
  }{2016a}]{ZhangWCLMSB16}
Zhang, S.; Wu, Y.; Che, T.; Lin, Z.; Memisevic, R.; Salakhutdinov, R.; and
  Bengio, Y.
\newblock 2016a.
\newblock Architectural complexity measures of recurrent neural networks.
\newblock {\em CoRR} abs/1602.08210.

\bibitem[\protect\citeauthoryear{Zhang \bgroup et al\mbox.\egroup
  }{2016b}]{zhang2016architectural}
Zhang, S.; Wu, Y.; Che, T.; Lin, Z.; Memisevic, R.; Salakhutdinov, R.~R.; and
  Bengio, Y.
\newblock 2016b.
\newblock Architectural complexity measures of recurrent neural networks.
\newblock In {\em Advances in Neural Information Processing Systems},
  1822--1830.

\end{thebibliography}
\bibliographystyle{aaai}
\end{document}